\documentclass[preprint,12pt,authoryear]{elsarticle}

\usepackage{amssymb}
\usepackage{graphicx}
\usepackage{amsmath}
\usepackage{amssymb}
\usepackage{booktabs}

\usepackage{changepage}
\usepackage{extramarks}
\usepackage{fancyhdr}
\usepackage{lastpage}
\usepackage{setspace}
\usepackage{soul}
\usepackage{xspace}
\usepackage{amssymb}

\usepackage{adjustbox}
\usepackage{array}
\usepackage{booktabs}
\usepackage{colortbl}
\usepackage{float,wrapfig}
\usepackage{hhline}
\usepackage{multirow}
\usepackage{subcaption} 
\usepackage{mathtools}

\usepackage{algorithm, algpseudocode}
\usepackage{enumerate}

\usepackage{listings}

\usepackage{enumitem}
\usepackage{graphics}

\usepackage[english]{babel}

\journal{Neural Networks}

\begin{document}

\begin{frontmatter}



\title{Divide-and-Conquer the NAS puzzle in Resource Constrained Federated Learning Systems}



\author[mymainaddress]{Yeshwanth Venkatesha\corref{mycorrespondingauthor}}
\ead{yeshwanth.venkatesha@yale.edu}
\author[mymainaddress]{Youngeun Kim}
\author[mymainaddress]{Hyoungseob Park}
\author[mymainaddress]{Priyadarshini Panda}



\cortext[mycorrespondingauthor]{Corresponding author}

\address[mymainaddress]{Department of  Electrical Engineering, 
Yale University, New Haven, Connecticut, USA}

\begin{abstract}
Federated Learning (FL) is a privacy-preserving distributed machine learning approach geared towards applications in edge devices. However, the problem of designing custom neural architectures in federated environments is not tackled from the perspective of overall system efficiency. In this paper, we propose DC-NAS---a divide-and-conquer approach that performs supernet-based Neural Architecture Search (NAS) in a federated system by systematically sampling the search space. We propose a novel diversified sampling strategy that balances exploration and exploitation of the search space by initially maximizing the distance between the samples and progressively shrinking this distance as the training progresses. We then perform channel pruning to reduce the training complexity at the devices further. We show that our approach outperforms several sampling strategies including Hadamard sampling, where the samples are maximally separated. We evaluate our method on the CIFAR10, CIFAR100, EMNIST, and TinyImagenet benchmarks and show a comprehensive analysis of different aspects of federated learning such as scalability, and non-IID data. DC-NAS achieves near iso-accuracy as compared to full-scale federated NAS with 50\% fewer resources.

\end{abstract}




\end{frontmatter}


\section{Introduction}
Deep learning has radically changed the landscape of AI with unprecedented success in computer vision, language processing, and several other domains such as recommendation engines and bioinformatics \cite{lecun2015deep, voulodimos2018deep, min2017deep, young2018recent, zhang2019deep}. 
Given that most of these applications are being deployed on millions of edge devices containing sensitive data, privacy-preserving distributed learning paradigms such as federated learning have garnered significant interest \cite{mcmahan2017communication, konevcny2016federated, chen2019deep, voghoei2018deep, vestias2020moving, li2018learning, kairouz2021advances}. 
A crucial part of the success of deep learning is attributed to the design of novel architectures \cite{simonyan2014very, he2016deep, szegedy2015going, szegedy2016rethinking, szegedy2017inception, liu2017survey}. Neural Architecture Search (NAS), the technique of searching for the optimal network architecture removes the tedious human effort of handcrafting architectures \cite{zoph2016neural, pham2018efficient, liu2018darts, liu2018progressive, tan2019mnasnet, elsken2019neural, wistuba2019survey}. With deep learning being rapidly applied in various domains, NAS is one of the promising steps toward democratizing AI.
Standard federated learning methods assume a fixed network architecture that is often manually designed beforehand for a given problem and hence might not be optimal. 
Further, the devices in practical federated systems are often constrained in their resources such as memory, storage, and processing power. Hence, it is prudent to develop practical methods to perform NAS in a federated learning system accounting for device resource constraints. 
While there is significant progress in compression techniques for reducing the complexity of neural networks \cite{neill2020overview, choi2020universal, blalock2020state, he2017channel, han2015deep, lin2017runtime, ashok2017n2n, luo2017thinet}, they do not perform NAS directly on the edge devices. It involves an additional step of first performing NAS offline followed by using the compression techniques to reduce the complexity of the model before deploying in the devices.

\begin{figure*}[t]
  \begin{center}
    \includegraphics[width=\textwidth]{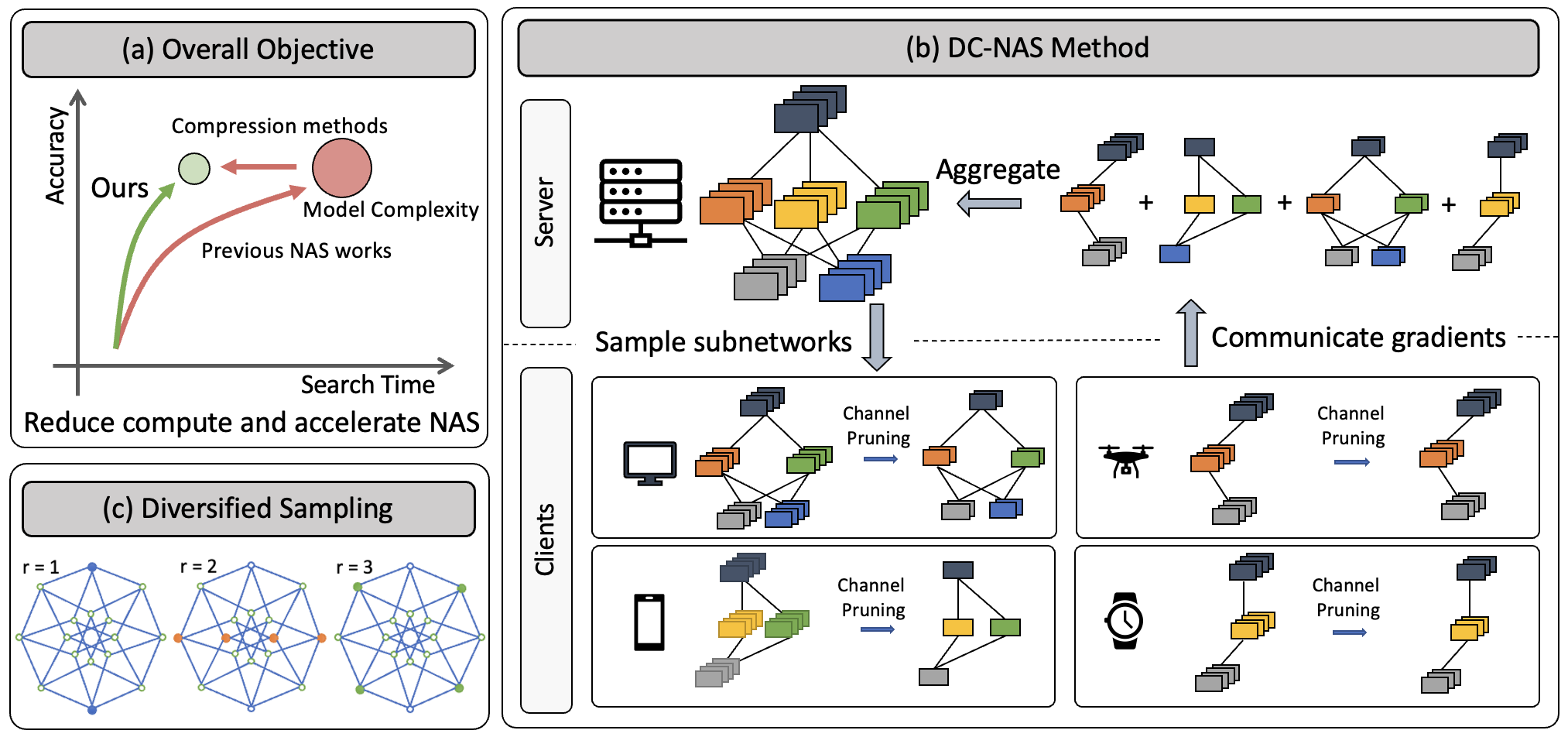}
  \end{center}
  \caption{Summary of DC-NAS approach. (a) Illustration of our objective of reducing the search time as well as the training complexity at the clients. (b) Illustration of the overall framework. The complexity of the model is reduced in two steps first by sampling the subnetworks from the server and then applying channel pruning at the clients. The communicated gradients after local training are aggregated and are used to update the supernet at the server. (c) Toy example showing our sampling approach on a 4-dimensional hypercube. Blue points picked in Round 1, orange in Round 2, and green in Round 3. } 
  \label{fig:main_method}
\end{figure*}

\begin{figure}[t]
  \begin{center}
    \includegraphics[width=\columnwidth]{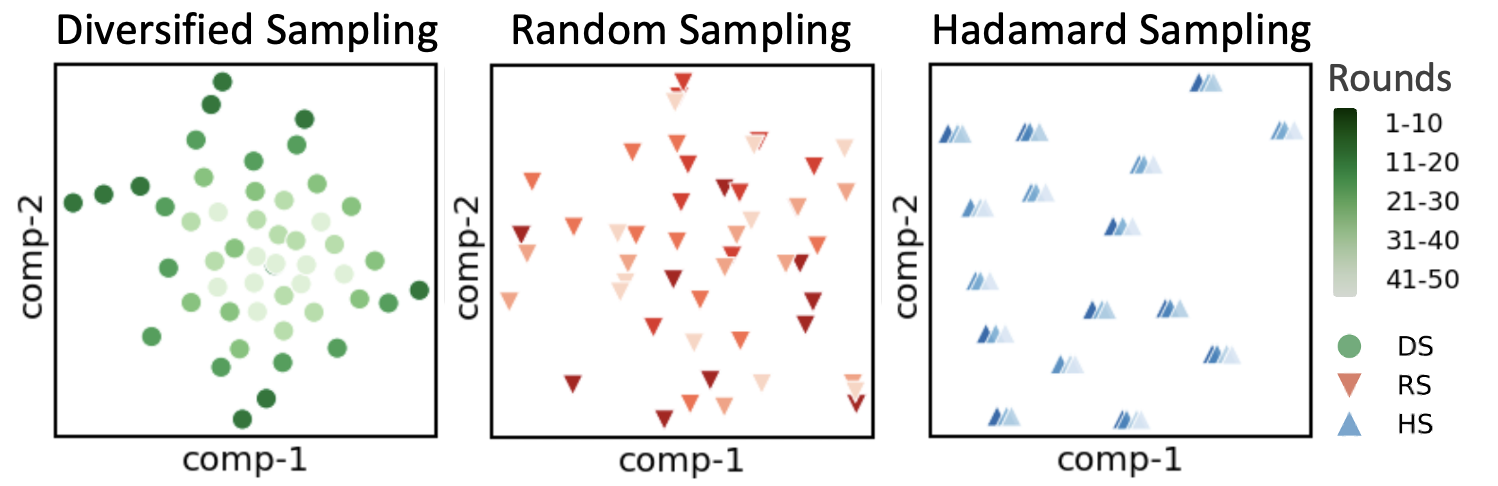}
  \end{center}
  \caption{t-SNE projections of the sampled points using diversified sampling, random sampling and Hadamard sampling. The sampled points follow a spiral pattern in diversified sampling showing the balance in exploration as well as exploitation. In contrast, random sampling and Hadamard sampling show the two extremes of exploration and exploitation. While random sampling is always in explore mode as seen by the arbitrary selection of points, Hadamard sampling is in full exploit mode by selecting the maximally separated points. Hence multiple points will overlap with each other when plotted on a 2D space. Note: The points in Hadamard sampling are perturbed to show that multiple samples are mapped to the same location.} 
  \label{fig:tsne_viz}
\end{figure}

In this paper, we propose DC-NAS---a Divide-and-Conquer approach to address the problem of NAS in a federated learning system consisting of resource-constrained devices. The objective of our method as illustrated in Fig. \ref{fig:main_method}a, is to accelerate the NAS process and at the same time reduce the training complexity of the models in the clients. 
To achieve this, we divide the search space among the clients and combine their updates analogous to assembling the pieces of a puzzle.
As shown in Figure \ref{fig:main_method}b, the model complexity on the clients is reduced in two stages. First, a smaller subnet is sampled for each client by pruning the connection topology of the supernet at the server. Then, within the client, the channels in the convolution layers are pruned to ensure that the resulting subnet fits within the client's resource constraints.
We then train the subnet using the clients' local data and aggregate the updates from different subnets into the larger supernet at the server. 
A straightforward method to sample the subnets is to randomly prune the connection topology until we achieve a sufficiently small network. However, this leads to an unstructured exploration of the search space and the aggregation of such subnets results in jagged updates consequently reducing the convergence rate similar to the observation in \cite{cai2019once}.
To address this issue, we propose a diversified sampling strategy (shown in Figure \ref{fig:main_method}c) that maps the problem of selecting subnets to sampling points on an $n$-dimensional hypercube. 
In the initial rounds, we maximize the distance between the samples and as the training progresses we reduce this distance thereby balancing the exploration of space as well as exploiting the known regions.
Visualized on a two-dimensional space using t-SNE \cite{van2008visualizing} in Fig. \ref{fig:tsne_viz}, this follows a spiral pattern by exploring the farther points in the initial rounds and progressively reducing the distance thereby exploiting the known regions in the later rounds. In contrast, random sampling selects the points arbitrarily. Finally in Hadamard sampling \cite{seberry2005some, horadam2012Hadamard, hedayat1978Hadamard, yarlagadda2012Hadamard}, while we obtain samples that are maximally separated in terms of hamming distance, the points are clustered when mapped on a 2D space showing that all the samples are rigidly interrelated hence restricting the exploration of the search space.
Our contributions can be summarized as follows:
\begin{itemize}
\item
We propose a divide-and-conquer approach to collaboratively perform neural architecture search in a federated environment with resource-constrained devices. We reduce the network complexity by pruning the connection topology followed by the channels to account for the compute budget in the client. 
\item
We propose a diversified sampling strategy to systematically sample subnets from the search space by sampling maximally separated samples in the initial rounds and progressively reducing the distance between the samples leading to an accelerated search. 
We show that our sampling process outperforms a spectrum of sampling strategies including Hadamard sampling which provides maximally separated samples.
\item
We show the effectiveness of our method with a comprehensive analysis on different aspects of federated learning such as non-IID data and scalability on CIFAR10, CIFAR100, EMNIST, and TinyImagenet datasets.
\end{itemize}


\section{Related Work}
Among several approaches towards tackling the problem of NAS, a majority are based on reinforcement learning (RL) \cite{baker2016designing, zoph2016neural, zoph2018learning, cai2018efficient, pham2018efficient}, evolutionary algorithms (EA) \cite{real2017large, suganuma2017genetic, liu2017hierarchical, real2019regularized, miikkulainen2019evolving, xie2017genetic, elsken2018efficient}, or gradient descent \cite{liu2018darts, xie2018snas, cai2018proxylessnas, he2020milenas, elsken2019neural, wistuba2019survey}. Among these, gradient-based methods offer the best trade-off between performance and resource consumption. 
Additionally, federated aggregation involves some form of averaging of the gradients from different clients which can be extended to architecture parameters, thus, making gradient-based NAS techniques a natural choice in federated environments.

NAS in federated environments has been previously approached from different perspectives.
FedNAS \cite{fednas} extends the DARTS \cite{liu2018darts} framework trained with mixed-level reformulation from MiLeNAS \cite{he2020milenas} to federated environments with a straightforward aggregation of both weights and architecture parameters. 
The authors of DecNAS \cite{xu2020federated} propose a technique to leverage the federated environment to conduct parallel training of multiple candidates for different constraints by extending NetAdapt \cite{yang2018netadapt} to federated environments.
The authors of RT-FedEvoNAS \cite{zhu2021real} take the approach of evolutionary algorithms to efficiently search the architecture space. 
The authors of \cite{cheng2022dpnas} and \cite{singh2020differentially} focus on the privacy aspect of federated NAS by adding noise to the gradients of architecture parameters and provide theoretical differential privacy guarantees. 
The authors of DFNAS \cite{garg2020direct} reduce the commonly used two-staged NAS approach (search and finetune) to a one-stage computationally lightweight method to obtain ready-to-deploy neural network models.
MGNAS \cite{pan2021privacy} provides a generic framework to NAS in federated learning by aggregating architectures from different clients by considering them as probabilistic graphs.
The authors of \cite{mushtaq2021spider} take the route of personalizing architecture design for each client whereas we define our problem as distributing the learning of a common model.
HANF \cite{seng2022hanf} proposes a method to perform both NAS and hyperparameter optimization using a combination of gradient-based and RL-based methods.
More recently, NAS in a federated environment is used in applications of IoT devices \cite{zhang2022towards} and secure medical data analysis \cite{liu2022federated}. 
While we limit our scope to horizontal federated learning where every participating client has the same feature set, there has been work on NAS for vertical federated learning where the clients hold feature-partitioned data \cite{liang2021self, liu2023cross}.
Finally, the authors of \cite{liu2021federated} and \cite{zhu2021federated} provide a broad overview of the problem of NAS in federated learning.

In contrast to previous approaches where NAS is performed independently on each client and the server is used to aggregate the models, DC-NAS performs NAS in a server-client integrated manner. The server distributes the work to every client by sampling a  different subnet and combines the knowledge learned at each client into the supernet.
While previous NAS approaches in federated learning train the supernet to find the best architecture, we train the entire supernet in a federated manner so that we can later sample architectures at different levels of complexity and deploy them off the shelf. This has the practical advantage of finding tailored architecture at different complexities based on the client's needs. 
While we use DARTS-based \cite{liu2018darts} search space in our experiments, the core sampling method can be extended to any NAS method that uses gradient or evolutionary techniques to train the architecture/weight parameters of a network.

\section{Methods}

\subsection{DC-NAS Framework}\label{section:dist_nas_framework}
The NAS search space is defined as a collection of weights and architecture parameters $(w, \alpha)$, where the set of weights $w$ represents the layers, and the architecture parameters $\alpha$ encode the connectivity among different layers \cite{liu2018darts}.
In a federated learning setup, we have a set of $C$ clients with a local dataset $D^{(c)}:{\{x_i, y_i\}}^{N^{(c)}}_{i = 1}$ at each client $c$ with $N^{(c)}$ samples out of total $N$ samples.
Given the set of weights $w$, architecture encoded by parameters $\alpha$ and a loss function $\mathcal{L}$, the global objective $f$ is optimized by minimizing the weighted sum of loss across all clients,
\begin{equation}
    \min_{w, \alpha} f(w, \alpha) = \min_{w, \alpha} \sum_{c = 1}^{C} \frac{N^{(c)}}{N}\cdot[\frac{1}{N^{(c)}}\sum_{i \in D^{(c)}} \mathcal{L}(x_i, y_i, w, \alpha)].
\end{equation}
Assuming that the client's resource constraints can be summarized to a scalar target sparsity $s^{(c)}$, we define  $s^{(c)}$ for each client $c$ that needs to be achieved to fit the subnet to the client's capability. To achieve this, we employ a two step process.
First, given the architecture of the supernet, the server samples a subnet for each client by pruning a fraction of connections. The subnet can be defined by a binary mask $m_{arch}^{(c)}$ with sparsity $s^{(c)}_{arch}$. Next, we apply channel pruning at the client with sparsity of $s^{(c)}_{channel}$ to further reduce the complexity of the subnet. 
We obtain the channel sparsity $s^{(c)}_{channel}$ required to achieve the overall target sparsity of $s^{(c)}$ as follows:
\begin{equation} \label{eqn:channel_pruning}
    s^{(c)}_{channel} = 1 - min(1, \frac{1 - s^{(c)}}{1 - s^{(c)}_{arch}}).
\end{equation}

\begin{figure}[t!]
\begin{algorithm}[H]
\caption{DC-NAS}\label{alg:framework}
\hspace*{\algorithmicindent} \textbf{Input:} Initial supernet: $(w_{0}, \alpha_{0})$; Number of local epochs: $E$; Number of rounds: $R$; $C$ clients indexed by $c$ each with local data $D^{(c)}$ \\
\hspace*{\algorithmicindent} \textbf{Output:} Trained supernet $(w^{*}, \alpha^{*})$

\begin{algorithmic}[1]
\State Initialize \textsc{Sampler} \Comment{(Ref. Algorithm \ref{alg:diversified-sampling})}
\For{round $r = 1, ..., R$}
    \State Get target sparsity $s^{(c)} = [s^{(1)},..., s^{(C)}]$ 
    \ForAll{clients $c = 1, ..., C$}
        \State $m_{arch}^{(c)} \gets \textsc{Sampler.SampleMask}()$ 
        \State $m_{channel}^{(c)} \gets [\{1\}^{s_{channel}^{(c)}n_c}, \{0\}^{(1 - s_{channel}^{(c)})n_c}]$ 
        \State $(w_r^{(c)}, \alpha_r^{(c)}) \gets m^{(c)} \otimes (w_r, \alpha_r)$ where $m^{(c)} = m_{arch}^{(c)} \circ m_{channel}^{(c)}$ \Comment{Apply mask} 
        \State $w_{r + 1}^{(c)}, \alpha_{r + 1}^{(c)} \gets \textsc{ LocalSearch}(w^{(c)}_{r}, \alpha_r^{(c)})$ 
    \EndFor
    \State $w_{r+1} \gets w_{r} + \sum_{c = 1}^{C} \frac{N^{(c)}}{N} \cdot m^{(c)} \otimes \nabla w^{(c)}_{r + 1} $
    \State $\alpha_{r+1} \gets \alpha_{r} + \sum_{c = 1}^{C}  \frac{N^{(c)}}{N} \cdot m^{(c)} \otimes \nabla \alpha_{r + 1}^{(c)} $ 
\EndFor

\Procedure {LocalSearch}{$w, \alpha$}
\For{epoch $e = 1, ..., E$}
\For{Minibatch in training and validation data in $D_c$}
\State $w \gets w - \eta_w \nabla_w \mathcal{L}_{tr}(w, \alpha)$
\State $\alpha \gets \alpha - \eta_{\alpha} ( \nabla_{\alpha} \mathcal{L}_{tr}(w, \alpha) + \lambda \nabla_{\alpha} \mathcal{L}_{val}(w, \alpha) )$
\EndFor
\EndFor 
\State \textbf{return} $(w, \alpha)$
\EndProcedure

\end{algorithmic}
\end{algorithm}
\end{figure}

As described in algorithm \ref{alg:framework}, for client $c$ in each round, we sample masks $m^{(c)}_{arch}$ for each client and prune the architecture of the supernet to obtain a sparse subnet. Then in each client, we keep only the first $s^{(c)}_{channel}$ fraction of the channels and prune out the rest. Let $m^{(c)}_{channel} = {\{0, 1\}}^{n_{c}}$ where $n_c$ is the number of channels, denote the binary mask of $n_c$ dimension with set bits indicating the active channels. With the notation of $m^{(c)} = m^{(c)}_{arch} \circ m^{(c)}_{channel}$ denoting the composition of the masks, we obtain the sparse subnet by applying $m^{(c)}$ over the weights and architecture parameters as $(w^{(c)}, \alpha^{(c)}) =  m^{(c)} \otimes (w, \alpha)$.
We then perform local search on each client over the subnet by alternating between updating $w^{(c)}$ and $\alpha^{(c)}$ for every minibatch of data \cite{he2020milenas, fednas}.
We then communicate the gradients of the trained subnets to the server. We update the corresponding weights and architecture parameters in the supernet by aggregating the gradients as,
\begin{equation}\label{eqn:agg_w}
    w = w + \sum_{c = 1}^{C} \frac{N^{(c)}}{N} \cdot m^{(c)} \otimes \nabla w^{(c)},
\end{equation}
\begin{equation}\label{eqn:agg_alpha}
    \alpha = \alpha + \sum_{c = 1}^{C} \frac{N^{(c)}}{N} \cdot m^{(c)} \otimes \nabla \alpha^{(c)}.
\end{equation}
Note that the updates are weighted according to the number of data samples ($N$) to reduce the variance of the updates \cite{mcmahan2017communication}.
At the end of the training procedure, we obtain a fully trained supernet from which high-performing subnets can be sampled.
While the subnets give a reasonable accuracy off the shelf, they can be further finetuned with the local client data to improve the accuracy. 
The strategy to sample the mask has a significant impact on the effectiveness of our framework. 
Here, we propose a diversified sampling strategy to systematically divide the search space among the clients.

\begin{figure}[ht]
\begin{algorithm}[H]
\caption{\textsc{Diversified Sampling - Sampler}}\label{alg:diversified-sampling}
\hspace*{\algorithmicindent} \textbf{Input:} Number of clients $C$ and number of rounds $R$.\\
\hspace*{\algorithmicindent} \textbf{Output:} Subnet masks $m_{arch}^{(c)}$ for client $c$ at round $r$. 

\begin{algorithmic}[1]
\State \texttt{parent\_nodes} $ \gets $ random sample 
\For{round $r = 1, ..., R$}
    \State \texttt{root} $\gets$ Pick a node from \texttt{parent\_nodes} 
    \State \texttt{offset = 1}
    \For{client $c = 1, ..., C$}
        
        \State $m_{arch}^{(c)}$ $\gets$ Flip every $r^{th}$ bit of \texttt{root} starting from \texttt{offset} bit
        \State \texttt{offset = offset + 1}
        \State \texttt{offset = offset \% r}
    \EndFor
    \State \texttt{parent\_nodes} $ \gets $ samples from current round $r$ 
\EndFor
\end{algorithmic}
\end{algorithm}
\end{figure}


\subsection{Diversified Sampling}\label{section:div-sampling}
Given a search space encoded by vector $\alpha$ of $n$ dimension, the binary mask $m_{arch}$ of the same dimension uniquely encodes all possible subnets. This results in a search space of $2^{n}$ possible architectures which can be visualized as points on an n-dimensional hypercube $Q_{n}$.
Given a node $u$ on $Q_n$, we can find the node that is $k$-hamming distance from it by flipping $k$ bits. 
In each round, we take the nodes sampled in the previous round and find the subsequent nodes that are at a certain distance from it. And we progressively reduce this distance as we go to later rounds. With this, we can explore the search space in a structured way with points sampled from a different region every round. 
To better understand the strategy consider the toy example of a 4-dimensional hypercube $Q_4$ shown in Fig. \ref{fig:main_method}c. 
Since all the nodes are symmetrical, without loss of generality we can assume we picked the first node. By flipping all the bits we reach the node that is diametrically opposite to the original node in the search space. In the next round, since we are flipping every alternate bit, we reach nodes that are at a hamming distance of $2$ from the nodes in the previous rounds. In subsequent rounds we flip every $3^{rd}, 4^{th}, 5^{th}$ bits and so on. At round $r$, this involves $r$ bit flips resulting in samples being $n/r$ hamming distance from the samples of previous rounds and $2n/r$ hamming distance among each other. Hence, we are systematically exploring the search space by sampling from the regions of the space that have not been previously explored. Additionally, given that the samples among the clients follow a pattern, the subnets fall into place like pieces of a puzzle upon aggregation. Note that this process can result in subnets that are beyond the capabilities of certain clients. For example, if we flip all the bits of a mask with a sparsity of 0.25, the resulting mask will have a sparsity of 0.75. To handle this we prune the channels of convolution layers to reach the required sparsity as described in Eqn. \ref{eqn:channel_pruning}.
We summarize our sampling method in algorithm \ref{alg:diversified-sampling}.



\section{Experiments}
\subsection{Experiment Setup}
We perform experiments with CIFAR10, CIFAR100 \cite{krizhevsky2009learning}, EMNIST \cite{cohen2017emnist} and TinyImagenet \cite{le2015tiny} datasets to show the effectiveness of our DC-NAS framework and diversified sampling. 
Following previous work \cite{fednas}, we partition the training set of CIFAR10 among 8 clients which is further divided into local training and validation sets. 
We simulate the non-IID scenario by sampling the proportions of classes using Dirichlet distribution with concentration parameter $\alpha_{iid} =0.5$ similar to previous work \cite{fednas, wang2020federated}.
The default hyperparameters that are listed in Table \ref{tab:hyperparams} are used across our experiments unless mentioned otherwise. We use a system with four V100 GPUs each with 32GB of GPU memory and an 18-core CPU with 8GB of memory per core.

\begin{table}[h]
  \caption{Default Hyperparameters.}
  \label{tab:hyperparams}
  \centering
  \begin{tabular}{lc|lc}
    \toprule
    Parameter     & Value   & Parameter     & Value \\
    \midrule
    \midrule
    Dataset & CIFAR10 &  Client Optimizer  &  Adam \\
    
    No. of Clients (C) & 8 & Grad clipping thr & 0.5 \\
    No. of rounds (R)  &  50 & Weight decay $w$ &  3e-4 \\
    Local Epochs (E) & 5 &  Weight decay $\alpha$ &  1e-3 \\
    Target Sparsity ($s^{(c)}$) & 0.5 &  LR for $w$ ($\eta_w$) & 0.001 \\
    Batch Size & 32 & LR for $\alpha$ ($\eta_{\alpha}$) & 3e-4 \\
    \bottomrule
  \end{tabular}
\end{table}

While we divide the training set among all the clients, we hold out the test set and use the accuracy achieved by the full supernet with composite operations 
on this global test set as the main performance metric across all our experiments. In addition to this, we also report the accuracy of the subnet formed by choosing only the operations with argmax($\alpha$). Further, this sampled network can be finetuned to improve accuracy. 
Finally, we track the training time and communication cost between the server and the clients to measure the computation and communication complexity by measuring training time and the model size respectively.




\subsection{Sampling Strategies}\label{results:ablation-study}
As explained in Section \ref{section:div-sampling}, our approach for exploring the search space involves finding samples that maximize the sum of the distance from previous samples in the initial rounds, and gradually reducing this distance to balance exploration and exploitation as the training progresses. In contrast, a simple random sampling method may not effectively cover the diverse regions of the space with a limited number of samples.
To study the effectiveness of our sampling method, we design a spectrum of sampling strategies by progressively increasing the interdependence between the samples. 

We start with the simple strategy of random sampling, which selects subnets for each client independently of each other. We then explore \textit{Antithetic sampling}, which reduces the variance of random samples by alternating the random selection between adjacent clients and using the complement of the selected subnet in the next client.
In the \textit{common sample} strategy, we randomly sample once per round and use this sample across all clients. This variant helps in understanding the effect of the interrelationship between the sampled subnets among the clients. \textit{Complement sample} is a special case of antithetic sampling where we generate a random sample only once per round and use it in half of the clients and use the complement of it in the remaining half. \textit{Hadamard sampling} uses Hadamard code to generate samples that are maximally separated from each other. 

In addition to our \textit{diversified sampling} described in section \ref{section:div-sampling}, we also define two variants, namely \textit{diversified sampling + reset} and \textit{diversified sampling + reset 10}. In \textit{diversified sampling + reset} we choose a random sample at the start of every round and perform the bit flip operation to obtain the subsequent samples. In \textit{diversified sampling + reset 10} we choose a random sample every 10 rounds. By comparing our method to these variants, we see that the interrelationship between the samples of the current round to the samples of the previous round has an impact on the search process. 
We use FedNAS \cite{fednas} as the \textit{baseline} which has no sampling and distributes/trains full supernet in every client.

\begin{figure}[t]
\centering

\subfloat[Comparison to random and Hadamard sampling.]{%
  \includegraphics[width=.45\columnwidth]{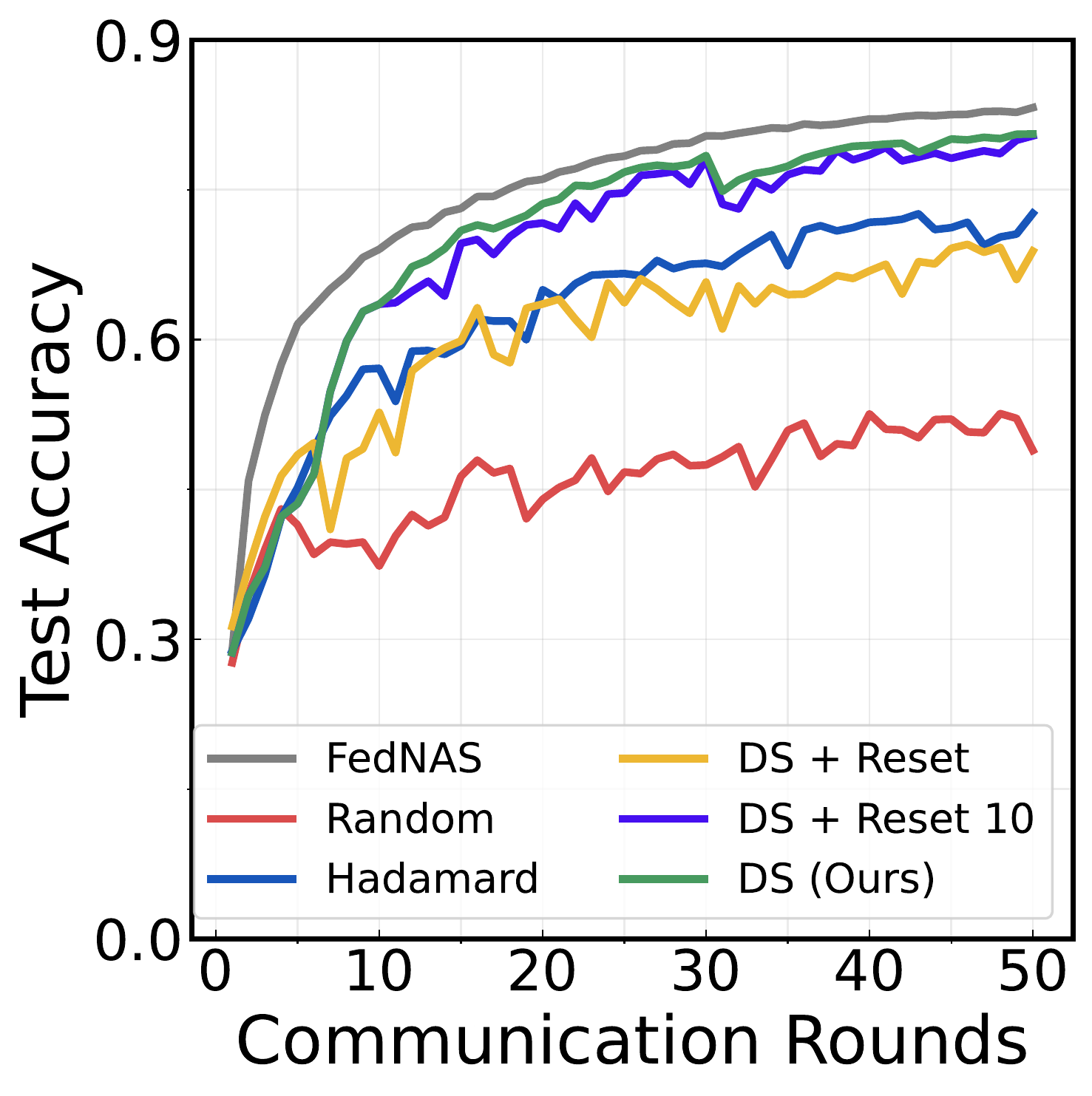}
  \label{subfig:variants1}}
\subfloat[Comparison of diversified sampling to additional sampling strategies.  ]{%
          \includegraphics[width=.45\columnwidth]{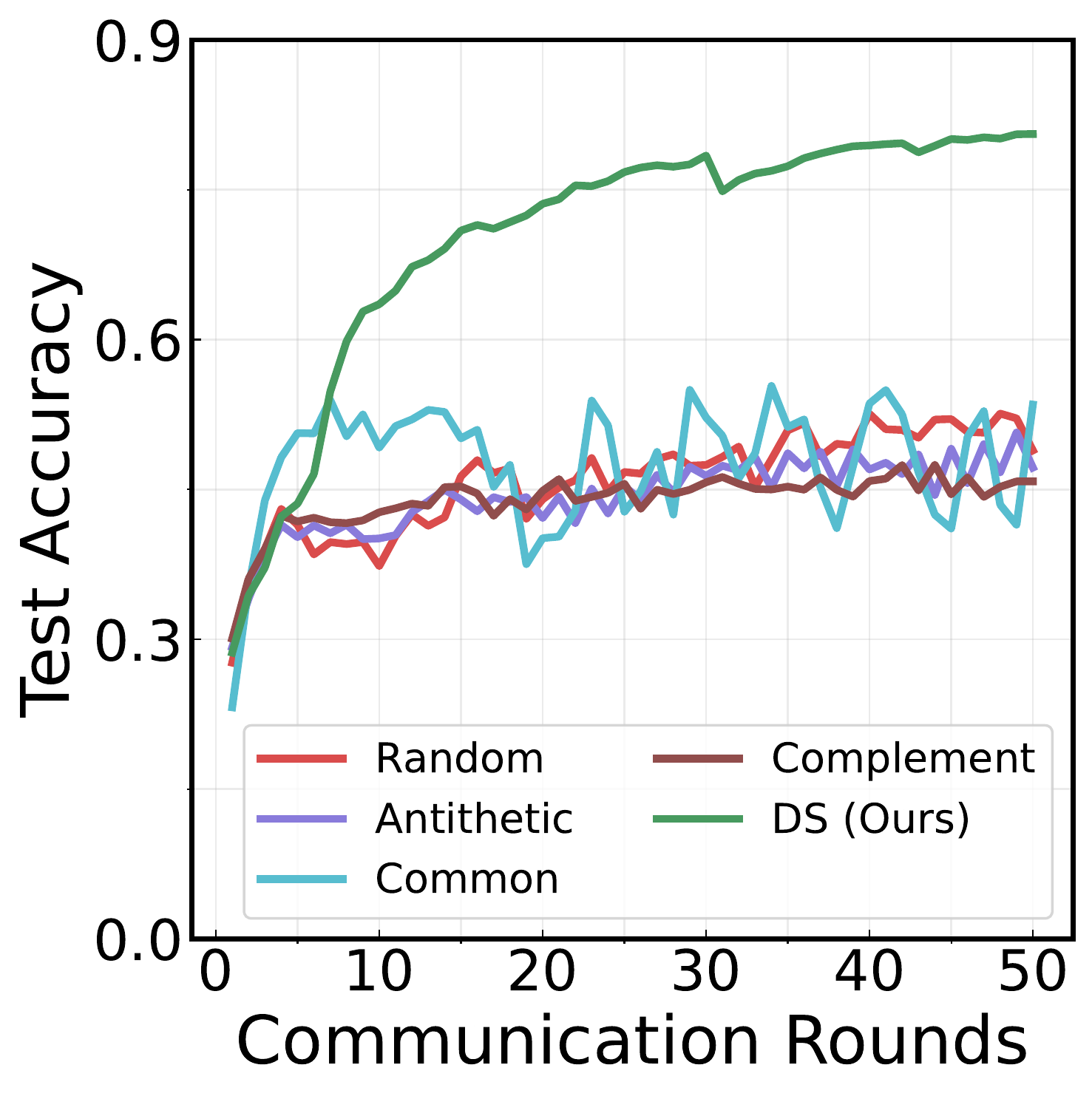}
          \label{subfig:variants2}}
\caption{ Performance of different sampling methods. (a) { The training progression across communication rounds of our diversified sampling as compared to random and Hadamard sampling.} (b) Performance comparison with additional variants including antithetic, complement sample and common sample.
    }
\label{fig:div-sampling-in-action}
\end{figure}

\begin{figure}[t!]
    \centering
    \begin{tabular}{cc}
    \adjustbox{valign=b}{\begin{tabular}{@{}c@{}}
    \subfloat[Distance between the samples across rounds. \label{subfig-2:interround}]{%
          \includegraphics[width=.4\linewidth]{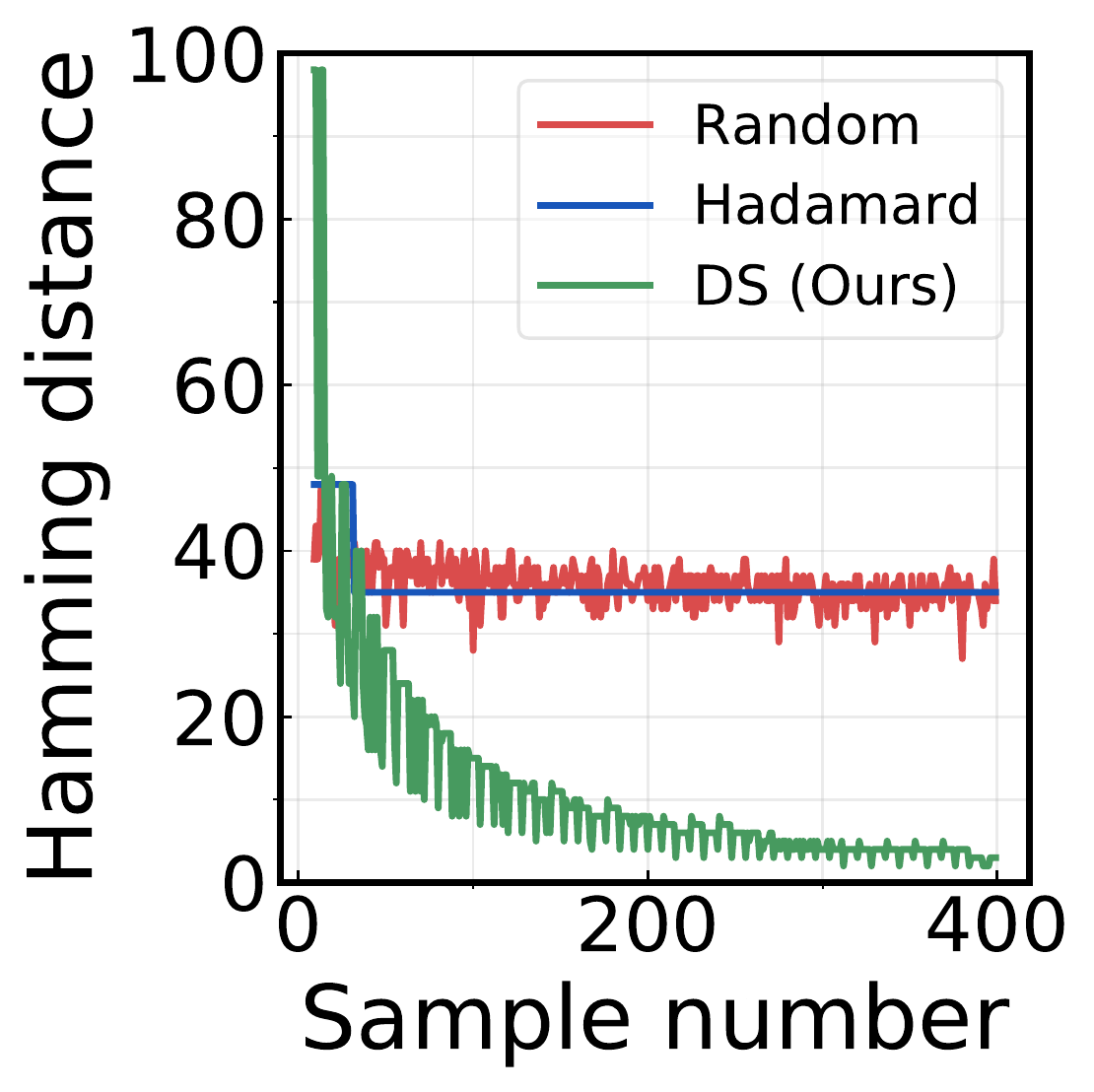}} \\
    \subfloat[Distance between the samples within rounds. \label{subfig-3:intraround}]{%
          \includegraphics[width=.4\linewidth]{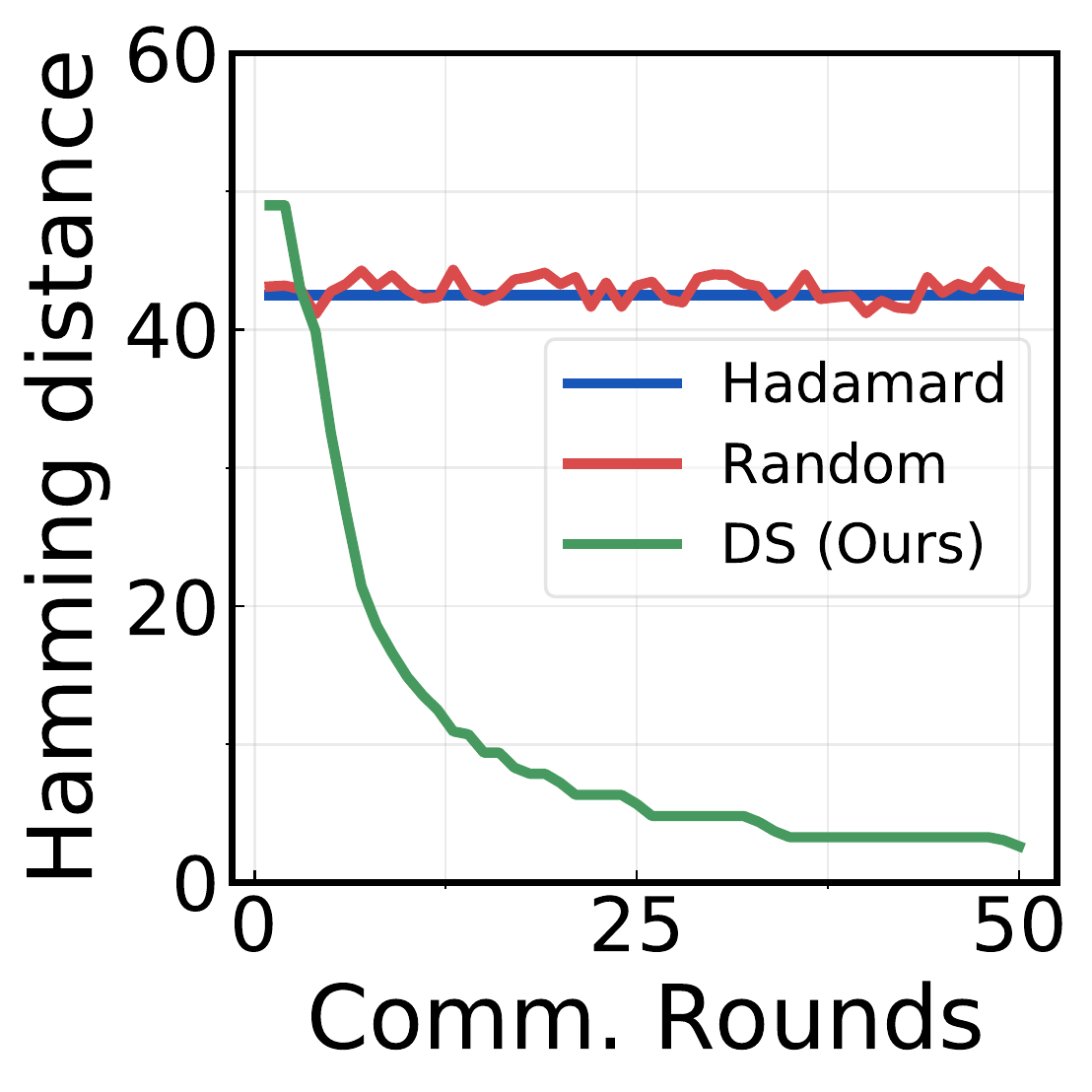}}
    \end{tabular}}
    
    \adjustbox{valign=b}{\begin{tabular}{@{}c@{}}
    \subfloat[Visualization of mask overlaps among clients across rounds. \label{subfig-2:visualization}]{%
          \includegraphics[width=.38\linewidth]{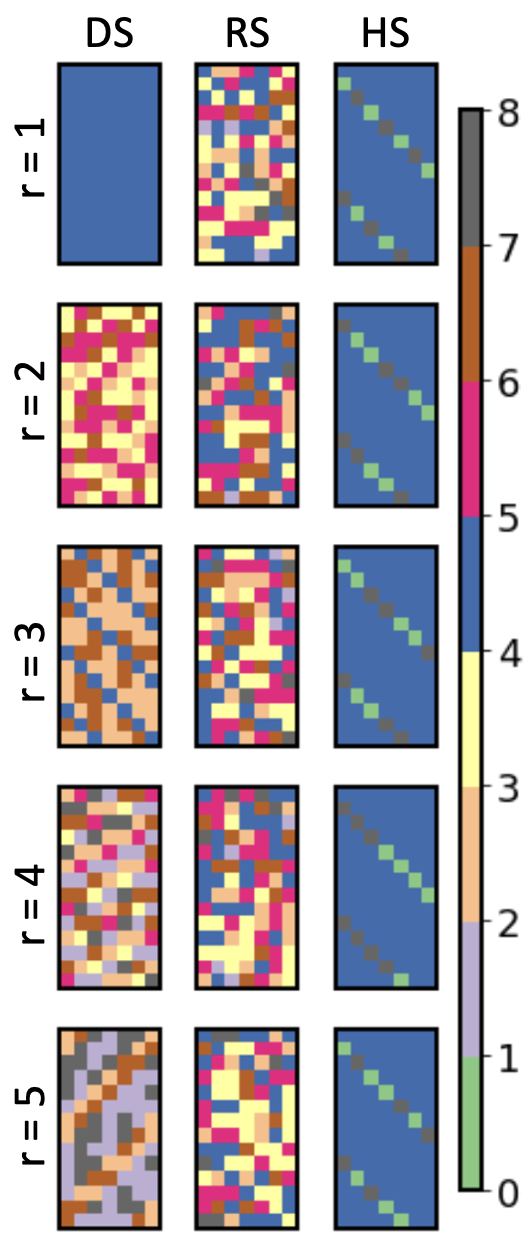}} \\
    
    \end{tabular}}

    \end{tabular}
    \caption{Diversified sampling in action. (a) Minimum hamming distance to all the previous samples across rounds. (b) Average hamming distance between all pairs of the samples at each round. (c) Visualization of distribution of operations among the clients across rounds for our diversified sampling compared with random and hadamard sampling. 
    }\label{fig:div-sampling-in-action-1}
  \end{figure}

While keeping all the parameters identical (from Table \ref{tab:hyperparams}), we vary only the sampling strategy and compare the test accuracy of the supernet across rounds in Fig. \ref{subfig:variants1} and Fig. \ref{subfig:variants2}. 
We observe that diversified sampling outperforms all the variants and achieves nearly the same performance as that of the FedNAS baseline. 
Note that in the beginning, all the variants show a similar trend because the interdependence of the samples is not yet pronounced as the total number of explored samples is low. Once we have explored the space sufficiently and the supernet is trained to a certain extent, the advantage of our structured exploration strategy with diversified sampling is evident with a significant performance improvement.

While antithetic sampling and complement sampling have no observable impact over random sampling, using common samples across all clients helps in the initial stages of the training owing to the absence of irregular overlaps in the subnets among the clients. However, as the training progresses, it becomes unstable due to high variance among the samples. 
Therefore the performance drop in random sampling, antithetic sampling, and complement sampling can be attributed to both irregular overlaps among the subnets of clients as well as inefficient exploration of the search space. 
Hadamard sampling is a strong baseline to compare our method as it yields samples that are proven to be maximally separated hence efficiently covering the search space. While it outperforms all the previous variants, it falls short to our diversified sampling indicating that the gradual reduction of exploration is effective in balancing finding new samples and exploiting the known areas.
We observe that diversified \textit{sampling+reset} with each round starting from a different random sample outperforms all the previous variants barring Hadamard sampling. This shows that reducing the distance between samples in the later rounds (by flipping every $r^{th}$ bit in round $r$) gives an observable improvement.
While we observe a performance drop when we reset every round, resetting every 10 rounds shows a minimal performance drop compared to the original diversified sampling. This reveals that the initial 10-15 rounds of exploration with diversified sampling are crucial.

\subsection{Diversified Sampling}
To understand the working of diversified sampling, we observe the hamming distances between the samples across the rounds as well as within the rounds. In Fig. \ref{subfig-2:interround}, at each sample, we plot the minimum hamming distance to all the previous samples. Note that this distance is high initially in diversified sampling and progressively reduces as the training progresses. This suggests aggressive exploration initially and then, a progressive reduction in exploration width. Similarly, the average distance among the samples within each round also follows the same trend (shown in Fig. \ref{subfig-3:intraround}). 
In Fig. \ref{subfig-2:visualization}, we plot the overlap of masks among the clients for the three variants. 
The value of each pixel denotes the number of clients where an operation is active. 

Note that in diversified sampling, initially the operations are distributed uniformly among all the clients (blue pixels showing each operation is active in 4 clients). In the later rounds, while some operations are exploited by having them active in more clients, certain operations are explored in some of the clients (as seen by grey and purple pixels respectively). On the other hand in RS, there is no such structure and HS follows a rigid structure of optimally distributing the operations among the clients in every round. 
This shows the balance in exploration and exploitation in diversified sampling as compared to complete exploration in random sampling and full exploitation in Hadamard sampling.

\begin{figure}[h]
\centering
\includegraphics[width=0.4\columnwidth]{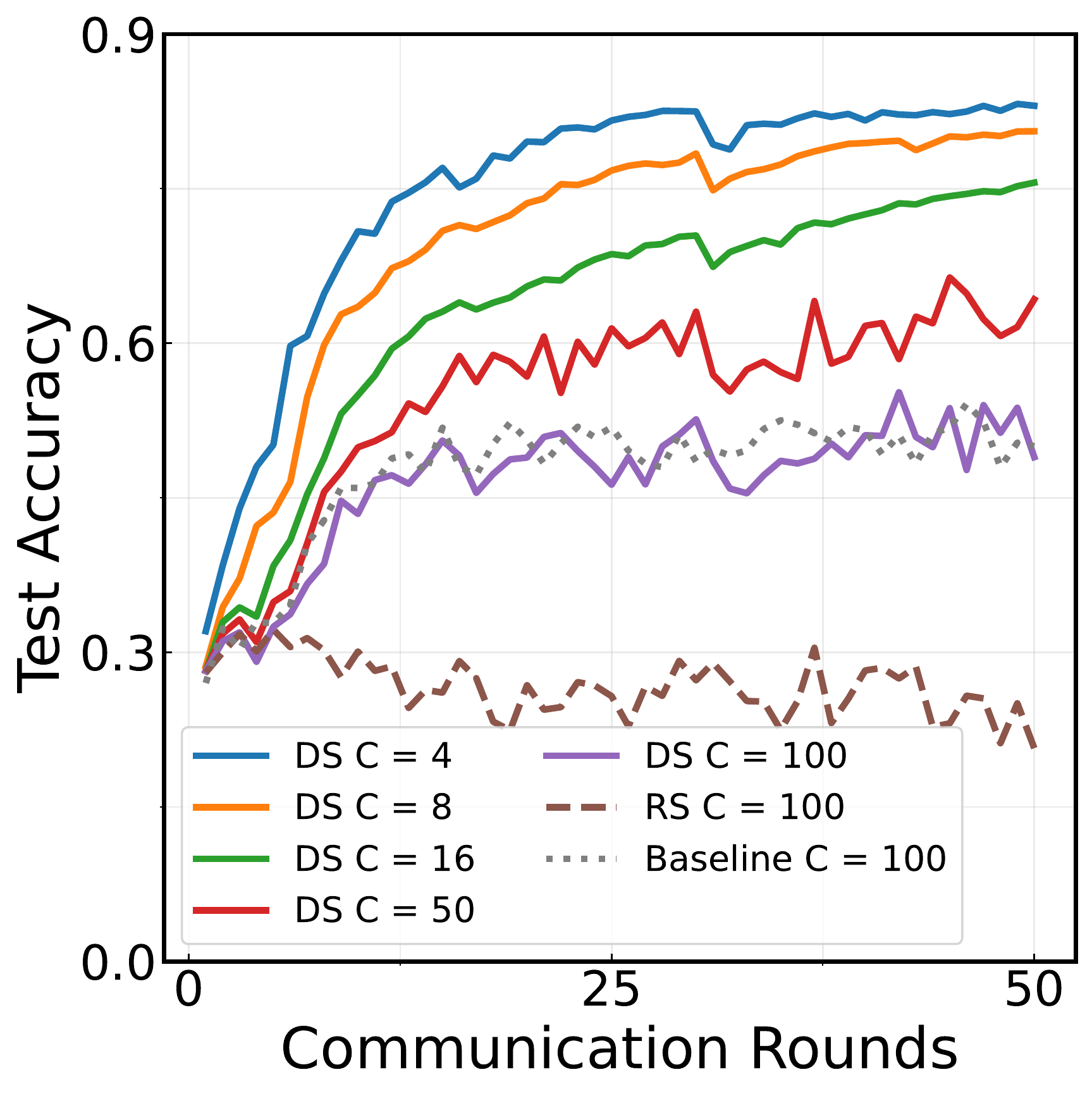}
\caption{Performance of diversified sampling at scale.}
\label{fig:scalability}
\end{figure}

\subsection{Scalability}\label{results:scale}
To evaluate the system's performance at scale, we increase the number of clients up to 100 in this experiment, as illustrated in Figure \ref{fig:scalability}. As expected, increasing the number of clients resulted in decreased overall performance and less stable training across all sampling strategies. This is a well-known phenomenon in standard federated learning, as the dataset size at each client reduces. For example, despite having a full supernet at every client, the FedNAS baseline performance decreased when the number of clients was 100. We notice a comparable pattern with our diversified sampling where the system's performance decreases as the number of clients increases. However, this is not caused by diversified sampling but rather by the characteristics of federated averaging (notably, the performance of DS was almost the same as the FedNAS baseline at 100 clients). On the other hand, due to the limited data samples at each client and the high variance introduced by random sampling, we notice a diverging behavior when training a 100-client system with random sampling.  
This experiment demonstrates that our approach can be scaled up to larger scenarios. Hence, our method can be applied to cross-silo as well as cross-device federated learning.

\begin{figure}[h]
\centering

\subfloat[CIFAR100]{%
  \includegraphics[clip,width=0.3\textwidth]{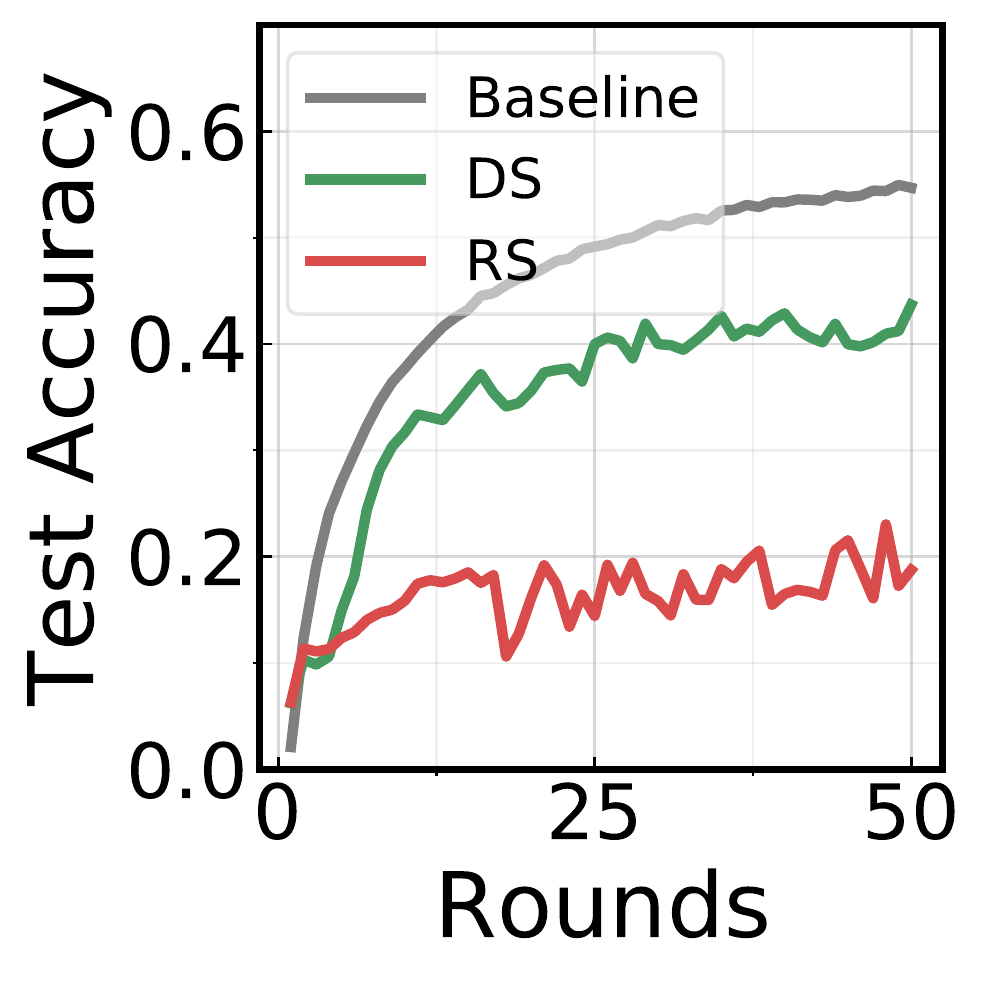}%
  \label{fig:CIFAR100}
}
\subfloat[EMNIST]{%
  \includegraphics[clip,width=0.3\textwidth]{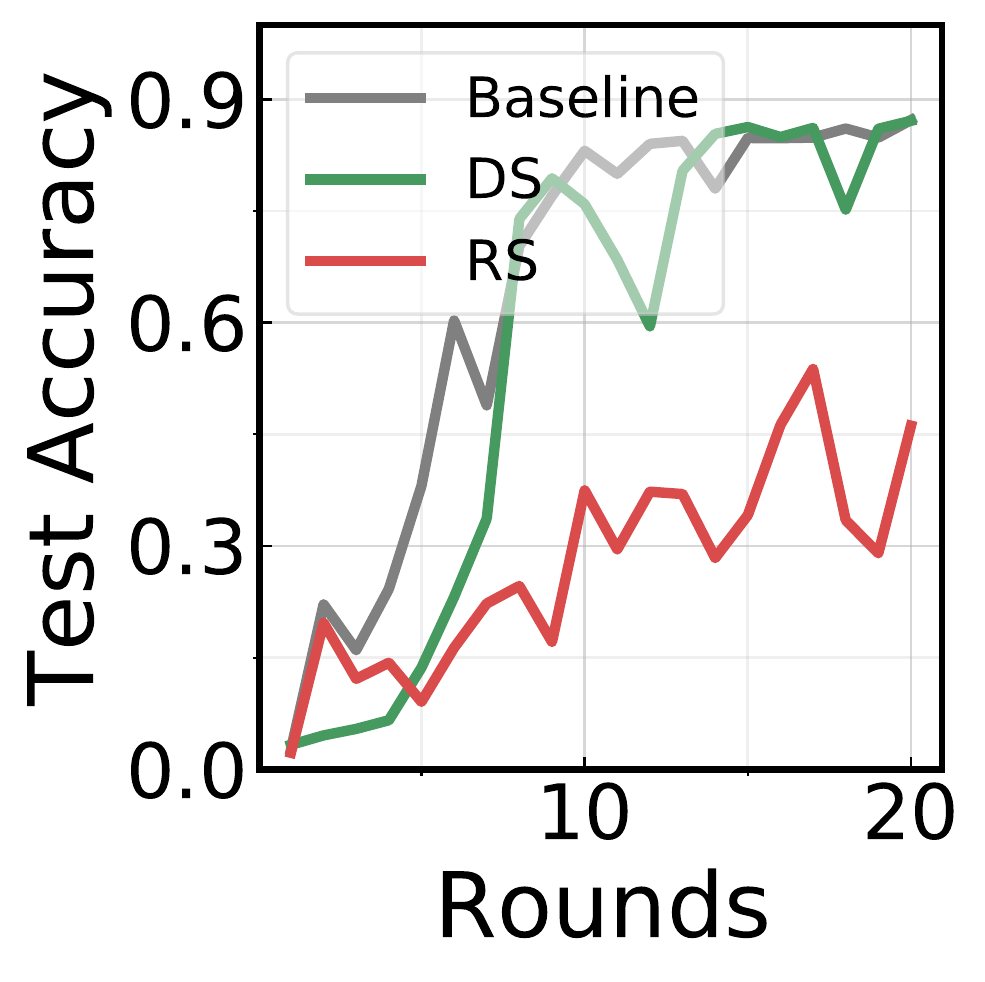}%
  \label{fig:EMNIST}
}
\subfloat[TinyImagenet]{%
  \includegraphics[clip,width=0.3\textwidth]{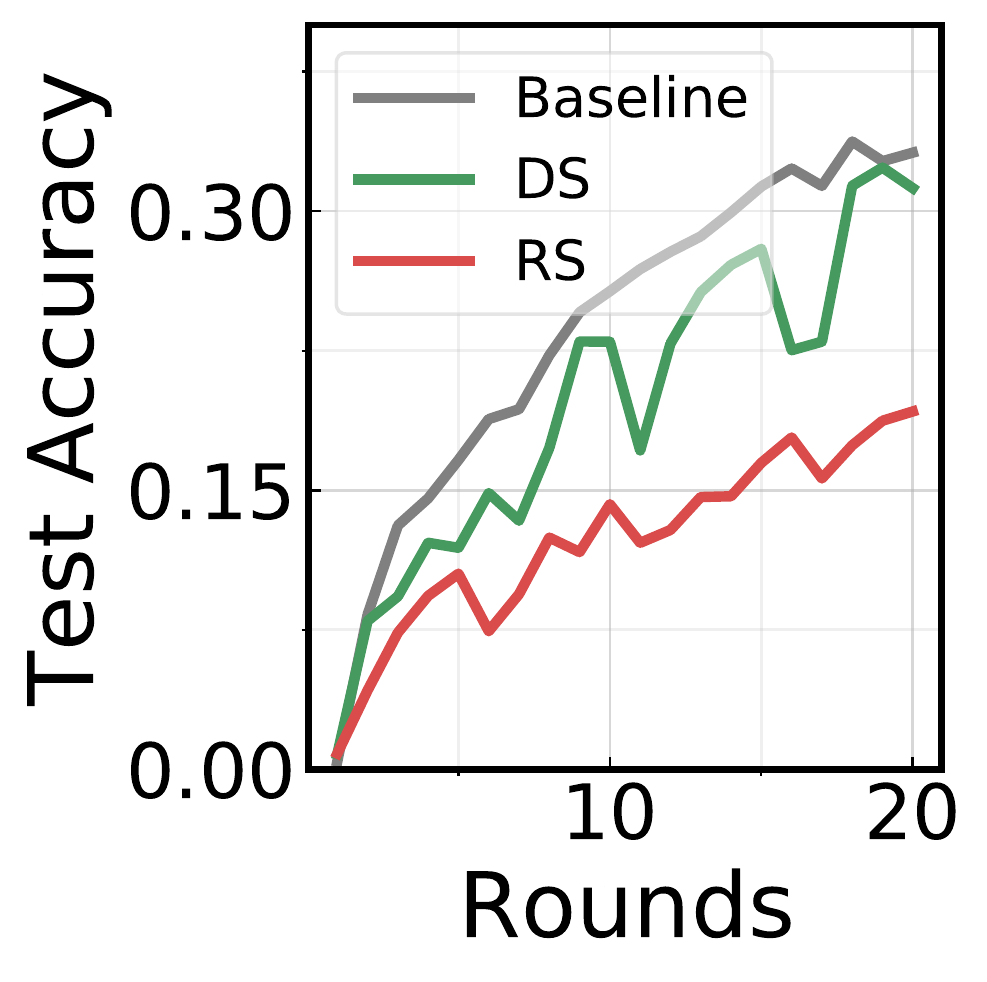}%
  \label{fig:TinyImagenet}
}
\caption{ Effectiveness of our method on additional datasets. }
\label{fig:other_datasets}
\end{figure}

\subsection{Additional Datasets}
CIFAR100, similar to CIFAR10 consists of $32 \times 32$ images with 100 classes. EMNIST consists of $28 \times 28$ dimensional grayscale images with 62 classes of handwritten alphabet. TinyImagenet is a trimmed down version of Imagenet and has 100000 images of 200 classes with image size reduced to $64 \times 64$. We divide all the datasets among 8 clients and use the same experimental setup. As shown in Fig. \ref{fig:other_datasets}, we observe that the trend across all datasets is similar to that of CIFAR10 with diversified sampling outperforming random sampling. While there is a performance drop as compared to the baseline in CIFAR100 and TinyImagenet due to the increased complexity of datasets, the performance gap between our method and random sampling is significant.

\begin{figure}[h]
\centering
\includegraphics[width=0.4\columnwidth]{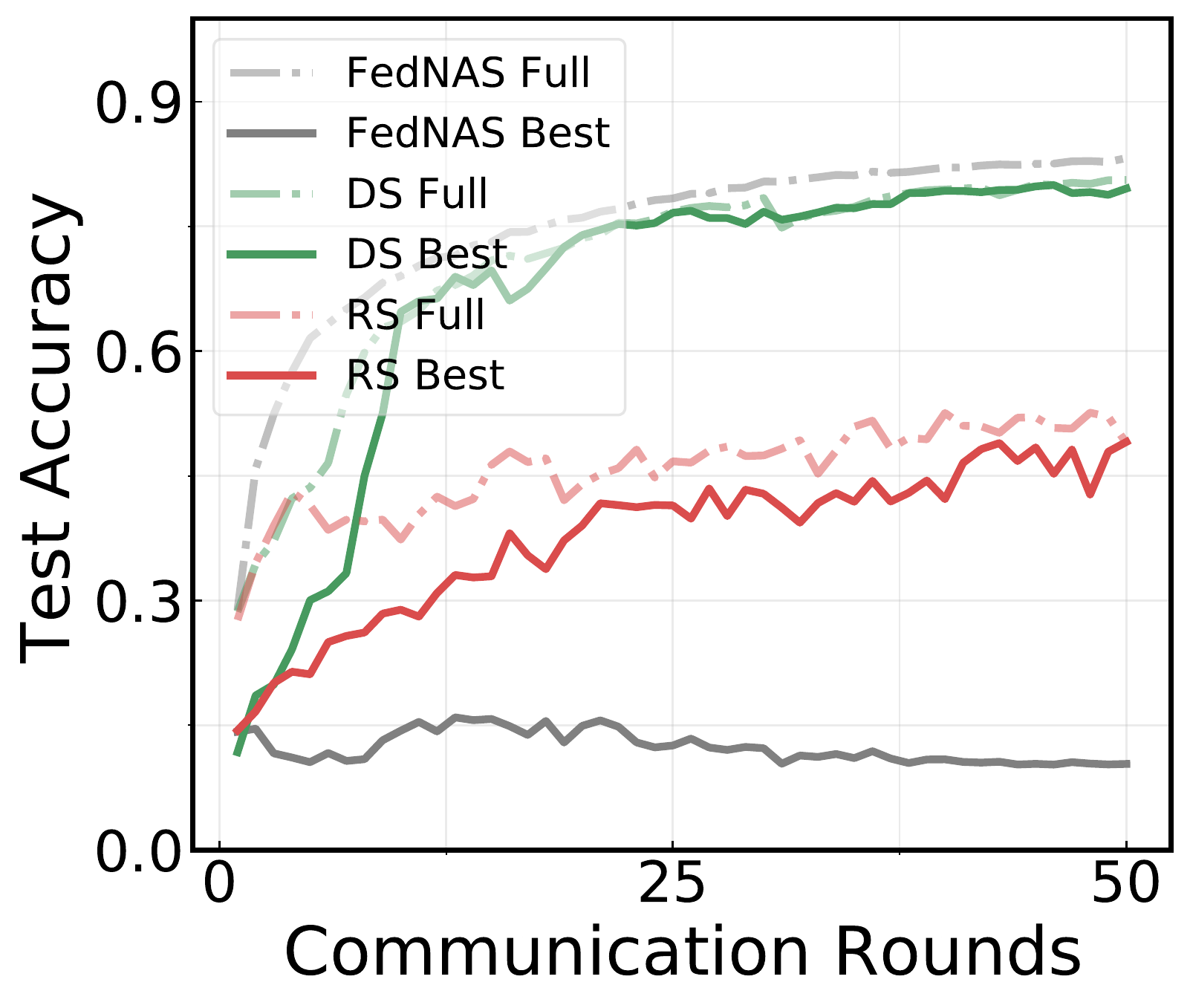}
\caption{Accuracy of the best subnet obtained by taking the argmax($\alpha$).}
\label{fig:argmax_acc}
\end{figure}

\subsection{Sampling Best Architecture}
In addition to the supernet, we also observe the accuracy of the best-performing subnet that is obtained by taking the argmax over the architecture parameters $\alpha$.
While it can be further finetuned, it is advantageous to have a subnet that can be deployed off the shelf. We compare the performance of diversified sampling and random sampling along with the FedNAS baseline in Fig. \ref{fig:argmax_acc}. We observe that by training with sampling, the best samples give near iso-performance off the shelf with that of the corresponding supernet. Still, diversified sampling yields no loss in accuracy between the best subnet and supernet.

In contrast for FedNAS baseline, when trained with full supernet, the best-performing architecture suffers a huge performance drop and needs to be finetuned before deploying.
This is because when trained with sampling, every time we are choosing a different part of the supernet making every component learn the underlying function independently. On the other hand, if full supernet is available at all times as with FedNAS, it learns as a single unit and, hence a sample from it fails to give reasonable accuracy without finetuning. This is analogous to training an ensemble of small models as compared to training one large model \cite{sagi2018ensemble}.

\begin{figure*}[t!]
\centering

\subfloat[Comparison of diversified sampling and random sampling at different $s^{(c)}$.]{%
  \includegraphics[clip,width=0.35\textwidth]{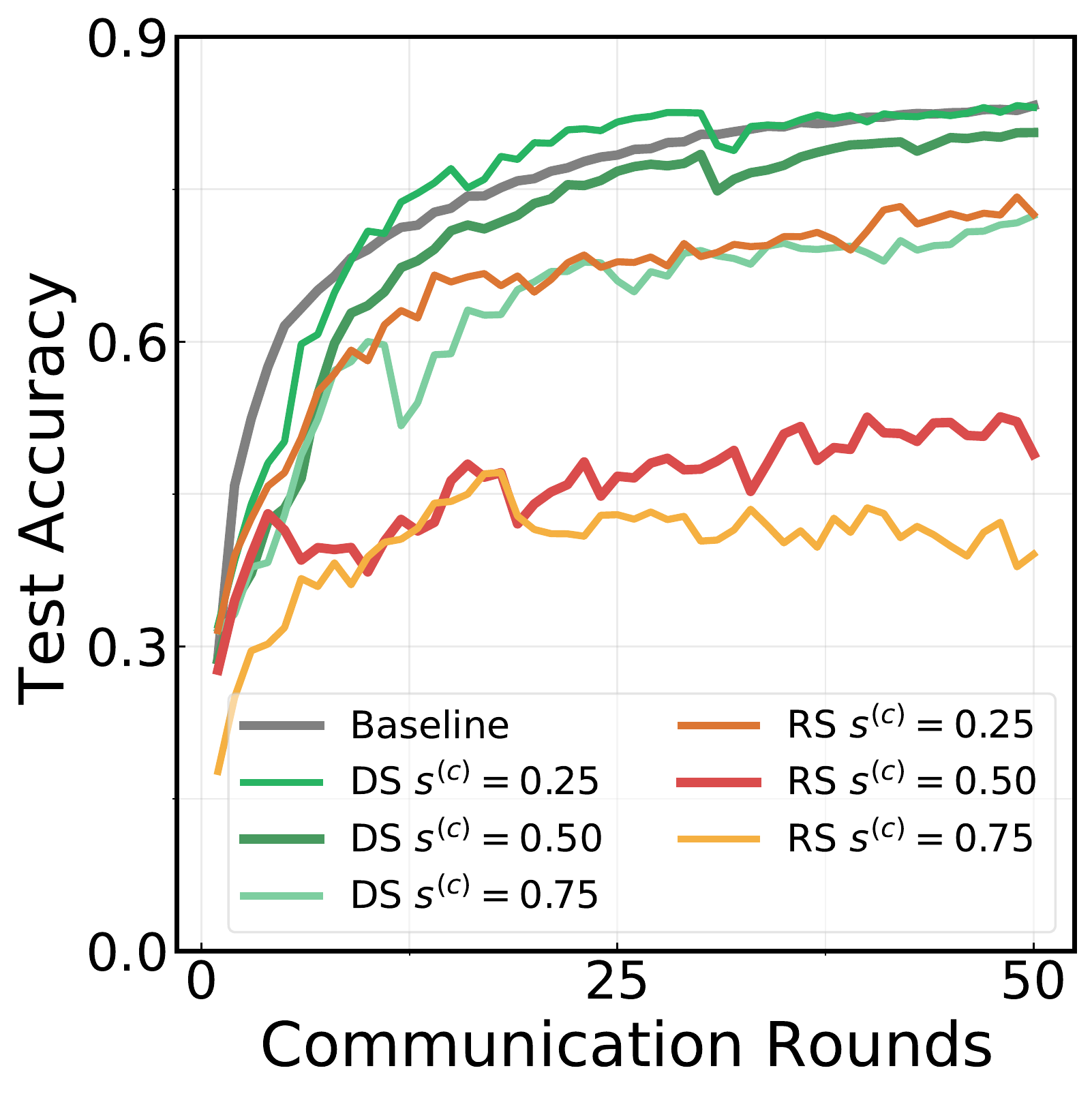}%
  \label{fig:homo_sparsity}
}
\quad
\subfloat[Comparison of training complexity.]{%
  \includegraphics[clip,width=0.24\textwidth]{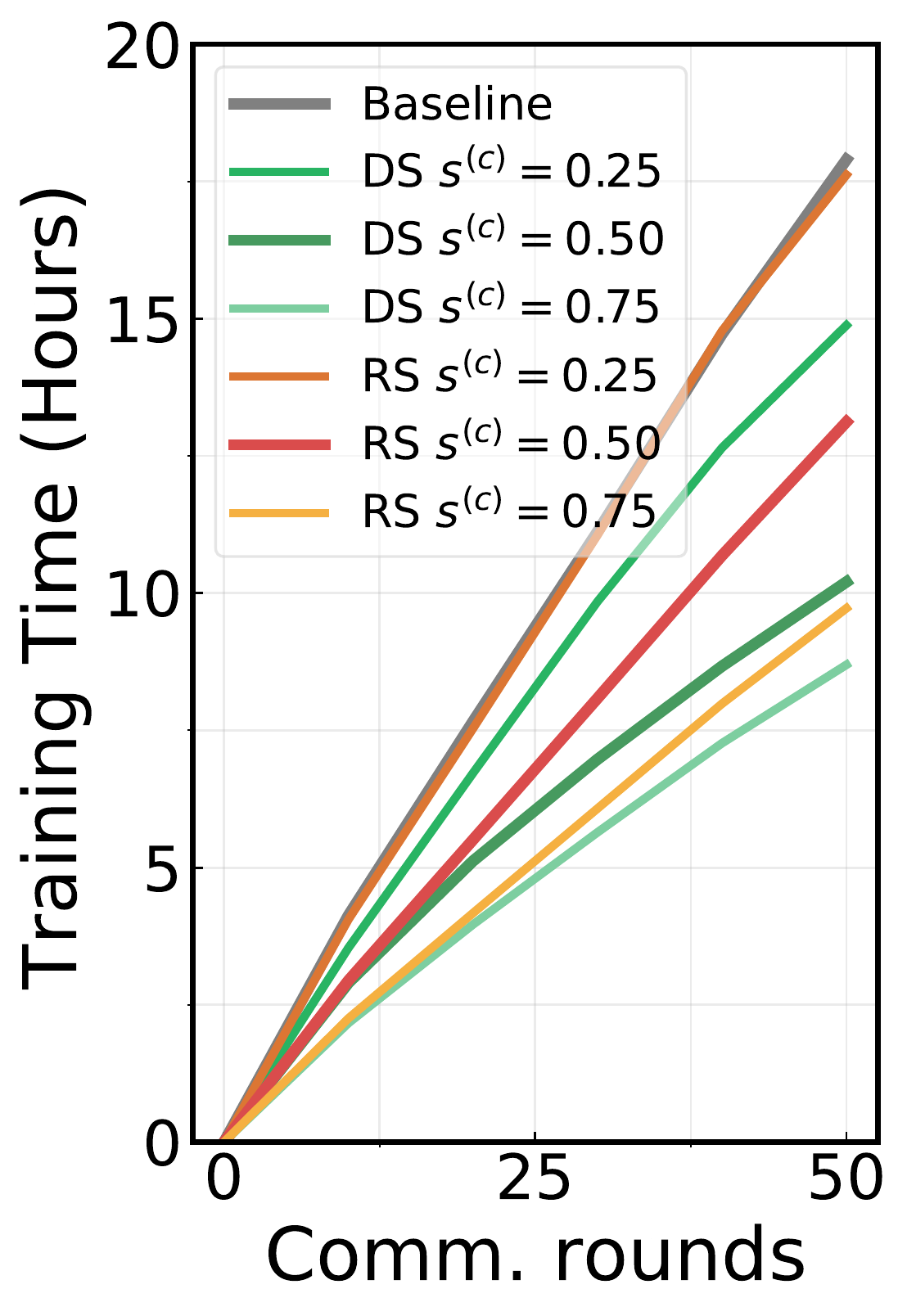}%
  \label{fig:train-time}
}
\quad
\subfloat[Comm. complexity of diversified sampling.]{%
  \includegraphics[clip,width=0.24\textwidth]{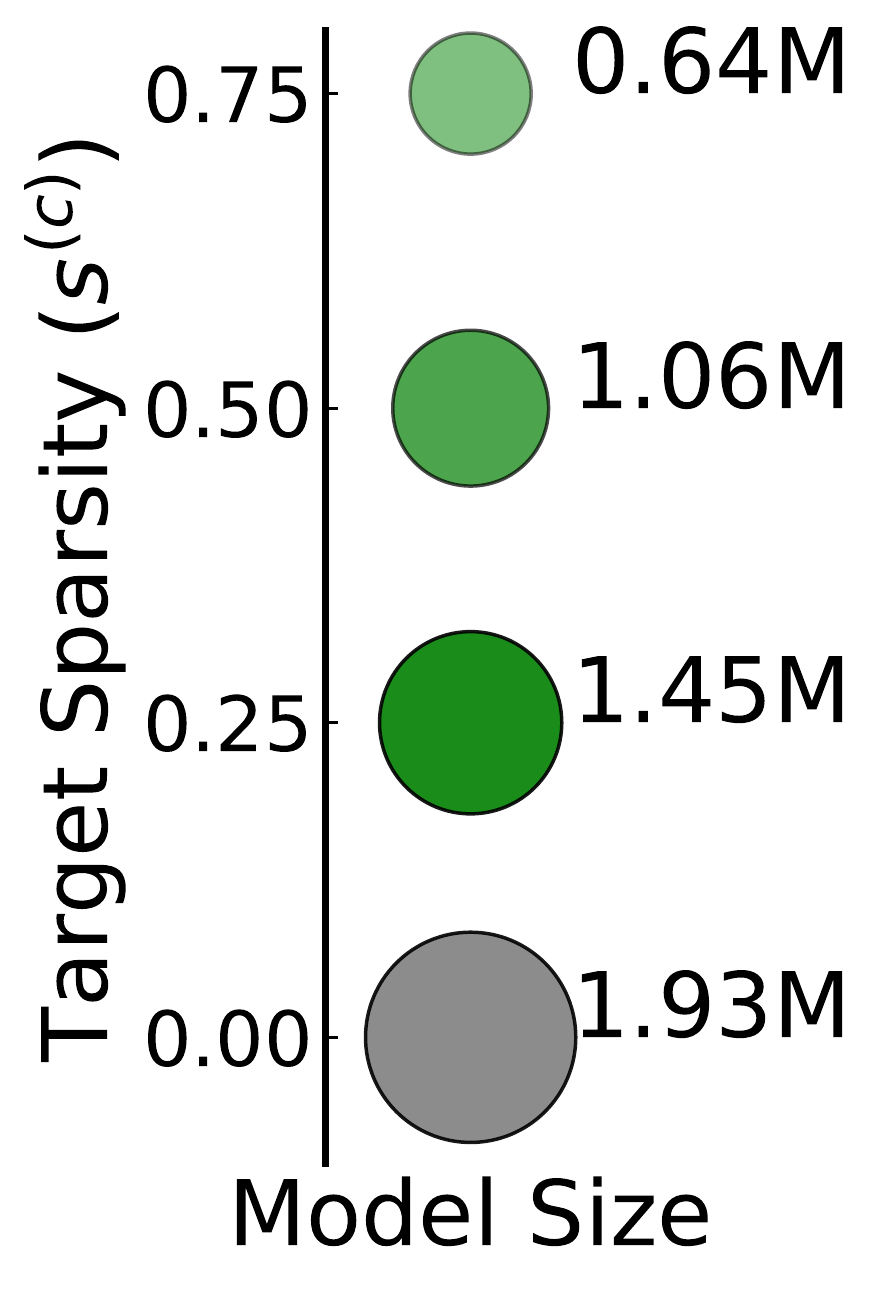}%
  \label{fig:comm_complexity}
}
\caption{Results showing the performance of diversified sampling (green) compared to random sampling (red) along with the FedNAS baseline (grey). (a) Shows the accuracy of the supernet at different values of $s^{(c)}$. (b) Shows the overall training time comparison of diversified sampling at different sparsity. (c) Communication complexity in terms of the average number of parameters in subnets. The communication complexity shown is for the FedNAS Baseline case (grey) with $s^{(c)}=0$ and Diversified sampling case (green) with different $s^{(c)}$.}
\end{figure*}

\begin{figure}[h]
\centering
\includegraphics[width=0.4\columnwidth]{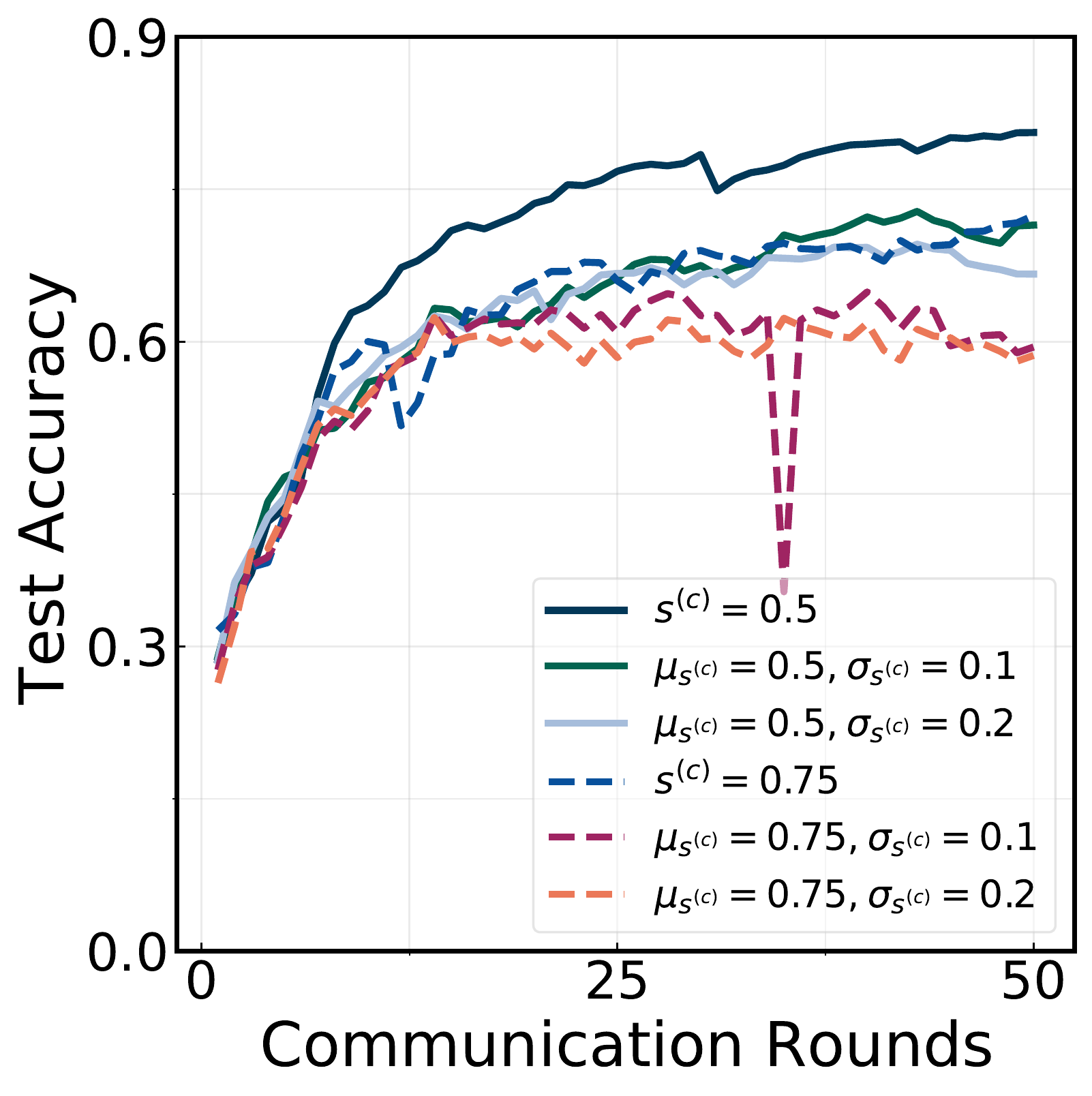}
\caption{Performance of diversified sampling with different sparsity constraints in each device. $\mu_s^{(c)}=0.5,  \sigma_s^{(c)} = 0.2$ denotes that clients can have heterogeneous sparsity $s^{(c)}$ in the range $[0.3, 0.7]$ with high probability.}
 \label{fig:hetero_study}
\end{figure}

\subsection{Architecture Sparsity and Search Complexity} \label{results:div-sampling}
One of the key parameters in our framework is target sparsity ($s^{(c)}$) as it constrains the complexity of the overall process. Here we measure it's effect on the accuracy, search time, and communication complexity.
In Fig. \ref{fig:homo_sparsity}, we use target sparsities of 0.25, 0.5, and 0.75 across all clients to compare diversified sampling and random sampling. Additionally, we show the FedNAS baseline case where there is no sampling which corresponds to $s^{(c)} = 0$.
We observe the test accuracy for the diversified sampling (DS) and random sampling (RS) with different values of $s^{(c)}$. We observe that increasing $s^{(c)}$ up to 0.5 has minimal effect on diversified sampling as the performance remains nearly the same as that of the baseline. Whereas, there is a significant drop in accuracy for the random sampling strategy which is only amplified with an increase in $s^{(c)}$. In the extreme case of $s^{(c)} = 0.75$, the test accuracy of random sampling drops to as low as 20\%. 

In Fig. \ref{fig:train-time} and Fig. \ref{fig:comm_complexity}, we measure the overall training time and the average number of parameters (model size) communicated in each round respectively. 
With different values of $s^{(c)}$, we observe a proportional reduction in training time and the volume of data transferred between the clients and the server.
Our method with $s^{(c)} = 0.5$ takes $\sim$10 hours for completing 50 rounds and achieves nearly the same accuracy as that of baseline which takes $\sim$18 hours. 
Note that while our method reduces the communication load as compared to baseline, the communication efficiency can be further improved by using state-of-the-art gradient compression techniques \cite{lin2017deep, albasyoni2020optimal}. 

In Fig. \ref{fig:hetero_study}, we simulate the case of heterogeneous clients by using different sparsities in different clients. In addition to uniform sparsity constraints of 50\% and 75\%, we simulate the variable sparsity constraint by sampling from a normal distribution at a standard deviation of 0.1 and 0.2.   While there is a slight drop in performance as compared to homogeneous sparsity, our method can handle the practical situations of clients having different constraints.

\begin{figure}[h]
\centering
  \includegraphics[clip,width=0.4\textwidth]{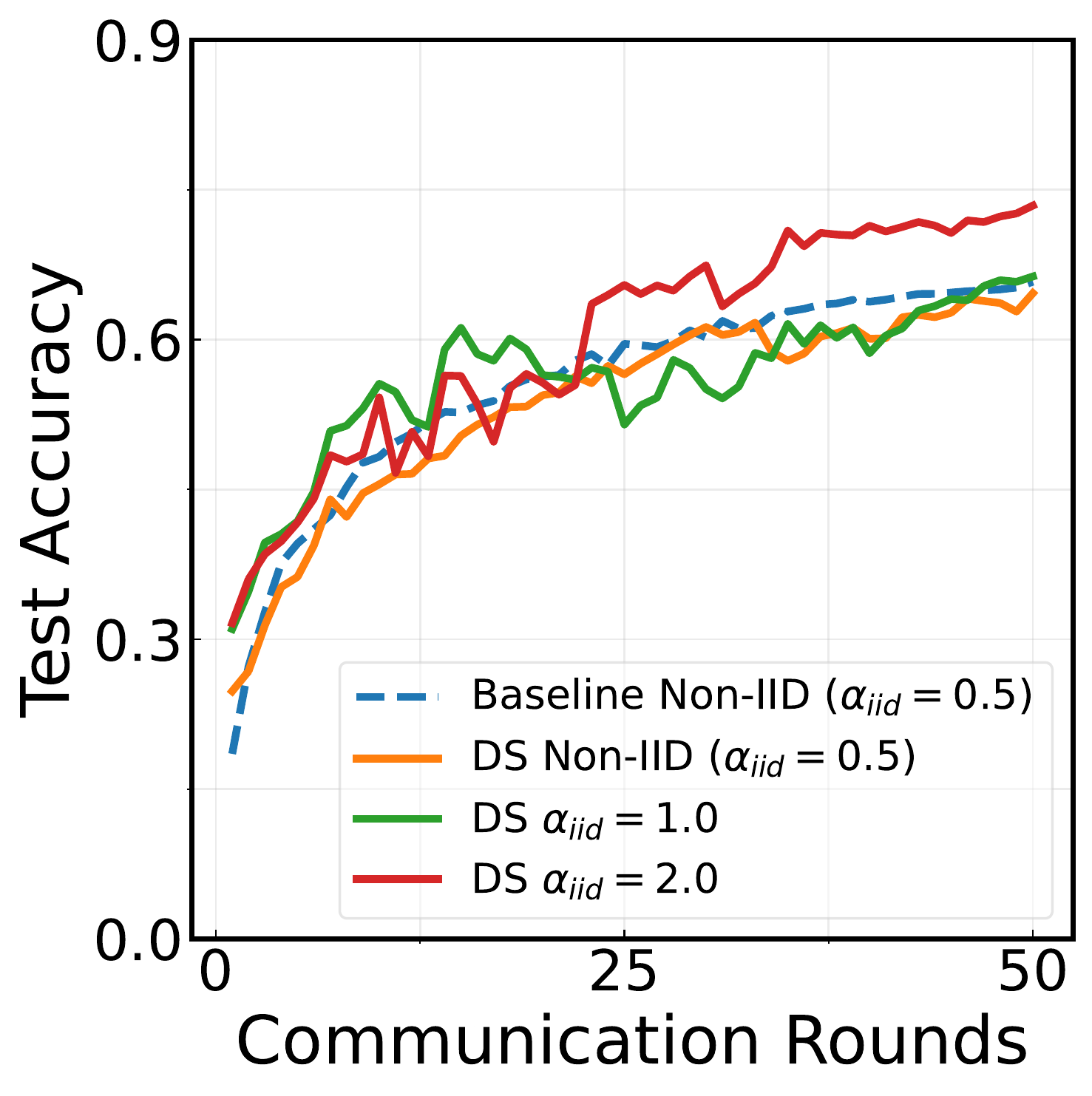}%
\caption{Performance of diversified sampling at different levels of skewness in the data. There is a performance drop as the data becomes more non-IID.}
\label{fig:non_iid_result}
\end{figure}

\subsection{Non-IID Partition} \label{results:non-iid}
In this section, we evaluate the effectiveness of our method when the data distribution among the clients is non-IID. 
We keep all parameters identical and vary the data distribution among clients from IID to various degrees of non-IID by tuning the alpha parameter in the Dirichlet distribution. 
{
In Fig. \ref{fig:non_iid_result}, we observe a drop in performance as the data becomes more non-IID i.e., as we decrease the $\alpha_{iid}$, the performance curve shifts downward. Note that the baseline also takes a similar hit in performance with non-IID data. Hence, we can attribute the performance drop to the challenge of handling non-IID data which is a well explored problem in federated learning. 
}

\subsection{Comparison to previous work} \label{results:prev-work}
We compare our results to the previous federated NAS works which have reported performance on CIFAR10 in Table \ref{tab:prev_work_comparison}. We also include the case of using only channel pruning without sampling (FedNAS + pruning) which serves as an ablation study to show the importance of subnet sampling. We include the case of fixed architecture (ResNet18) and its corresponding pruned version for completeness.
Since each of the previous methods have a different experimental setup, we quote the reported final accuracy reached by every method after finetuning along with the number of parameters, MAC operations, and communication rounds. Since with our divide and conquer approach, each client is training a smaller subnet, we reduce the number of MAC operations as well as the number of parameters to communicate between the clients and the server. Hence, we achieve near state-of-the-art accuracy with significantly less compute and communication complexity as compared to the state-of-the-art. 







\begin{table}[h!]
  \caption{Comparison to previous work. We quote the final accuracy, model size, number of MACs and number of communication rounds in the search phase as well as finetune(FT) phase. reported by the previous methods. We compare the communication complexity with number of parameters and compute complexity with number of MAC operations. Note that since SPIDER \cite{mushtaq2021spider} designs personalized architecture for each client, the reported metrics are averaged over the metrics obtained by the clients. }
  \label{tab:prev_work_comparison}
  \centering
  \resizebox{\columnwidth}{!}{%
  \begin{tabular}{lcccc}
    \toprule
    \multirow{2}{*}{Method} & \multirow{2}{*}{Acc (\%)} & \multirow{2}{*}{Params} & \multirow{2}{*}{MACs} & Rounds \\
     &  &  &  & (Search + FT) \\ 
    \midrule
    \midrule
    FedNAS \cite{fednas} & \multirow{1}{*}{91.43} & \multirow{1}{*}{1.93M} & \multirow{1}{*}{317.57M} & \multirow{1}{*}{50 + 50} \\
    
    {FedNAS + Pruning} & \multirow{1}{*}{79.90} & \multirow{1}{*}{0.97M} & \multirow{1}{*}{160.48M} & \multirow{1}{*}{50 + 50} \\
    
    \hline
    DFNAS \cite{dfnas} & \multirow{1}{*}{\textbf{92.11}} & \multirow{1}{*}{2.1M} & \multirow{1}{*}{-} & \multirow{1}{*}{150 + 0} \\
    
    RT-FedEvoNAS \cite{zhu2021real} & \multirow{1}{*}{86.68} & \multirow{1}{*}{-} & \multirow{1}{*}{279.60M} & \multirow{1}{*}{500 + 0} \\
    
    DPNAS \cite{cheng2022dpnas} & \multirow{1}{*}{68.33} & \multirow{1}{*}{0.53M} & \multirow{1}{*}{-} & \multirow{1}{*}{100 + 0} \\
    
    \multirow{1}{*}{DP-NAS \cite{singh2020differentially}} & \multirow{1}{*}{86.0} & \multirow{1}{*}{3.36M} & \multirow{1}{*}{-} & \multirow{1}{*}{50 + 50} \\

    MGNAS \cite{pan2021privacy} & \multirow{1}{*}{85.33} & \multirow{1}{*}{-} & \multirow{1}{*}{-} & \multirow{1}{*}{400 + 0} \\
    
    SPIDER \cite{mushtaq2021spider} & \multirow{1}{*}{92.0} & \multirow{1}{*}{345K} & \multirow{1}{*}{62M} & \multirow{1}{*}{1500 + 0} \\
    \multirow{1}{*}{HANF \cite{seng2022hanf}} & \multirow{1}{*}{90.0} & \multirow{1}{*}{-} & \multirow{1}{*}{-} & \multirow{1}{*}{120 + 1500} \\
    
    \hline
    \multirow{1}{*}{ResNet18} & \multirow{1}{*}{91.01} & \multirow{1}{*}{11.17M} & \multirow{1}{*}{556.65M} & \multirow{1}{*}{0 + 50} \\
    
    ResNet18 + Pruning & \multirow{1}{*}{85.42} & \multirow{1}{*}{5.81M} & \multirow{1}{*}{289.52M} & \multirow{1}{*}{0 + 50} \\
    \hline
    \multirow{1}{*}{\textbf{DC-NAS (Ours)}} & \multirow{1}{*}{90.21} & \multirow{1}{*}{\textbf{1.06M}} & \multirow{1}{*}{\textbf{186.64M}} & \multirow{1}{*}{50 + 50} \\

    \bottomrule
  \end{tabular}
  }
\end{table}

\section{Conclusion and Future Work}\label{section:conclusion}
We present a divide-and-conquer approach to perform NAS in a resource-constrained federated learning system efficiently. We propose a novel diversified sampling technique to accelerate the search process while being mindful of the compute capabilities of the clients. We show the effectiveness of our method over multiple datasets with experiments spanning different aspects of federated learning. 
While our work focuses on improving the efficiency of the search, there is room for improving the overall efficiency of the system.
For example, standard pruning and quantization can be used in addition to our approach to compress the final architecture before deploying.
During sampling the mask, we are disregarding the exact compute load and assuming that every operation is going to take similar compute resources. Hence, it is possible to extract further efficiency by explicitly weighting each operation proportional to the compute load of the operation. 
{Further, we provide a generic solution but the specifics of devices can have different implications on final training and communication complexity.}
In this paper, we show the effectiveness of our strategy empirically and leave the theoretical analysis to future work. 
{
While we show the scalability with respect to the number of clients with a preliminary experiment, it is worthwhile to study how it scales to a larger search space and larger datasets. 
Further, while we use DARTS search space in our experiments, the core method is not specific to DARTS and can be extended to any search space that can be modeled as a binary vector.}
Finally, since our method relies on the server to orchestrate the whole process, it is heavily dependent on the server's reliability. Hence, a decentralized peer-to-peer version of this problem is an interesting avenue to explore.



\section*{Acknowledgements}
This work was supported in part by CoCoSys, a JUMP2.0 center sponsored by DARPA and SRC, Google Research Scholar Award, the National Science Foundation CAREER Award, TII (Abu Dhabi), the DARPA AI Exploration (AIE) program, and the DoE MMICC center SEA-CROGS (Award \#DE-SC0023198).



\bibliographystyle{elsarticle-harv} 
\bibliography{egbib}




\end{document}